# A NERF-BASED COLOR CONSISTENCY METHOD FOR REMOTE SENSING IMAGES

*Zongcheng Zuo, Yuanxiang Li, Tongtong Zhang*

School of Aeronautics and Astronautics, Shanghai Jiao Tong University, Shanghai, China

## ABSTRACT

Due to different seasons, illumination, and atmospheric conditions, the photometric of the acquired image varies greatly, which leads to obvious stitching seams at the edges of the mosaic image. Traditional methods can be divided into two categories, one is absolute radiation correction and the other is relative radiation normalization. We propose a NeRF-based method of color consistency correction for multi-view images, which weaves image features together using implicit expressions, and then re-illuminates feature space to generate a fusion image with a new perspective. We chose Superview-1 satellite images and UAV images with large range and time difference for the experiment. Experimental results show that the synthesize image generated by our method has excellent visual effect and smooth color transition at the edges.

## 1. INTRODUCTION

Orthophoto is one of the important achievements of satellite imagery, which plays an important role in surveying and mapping, land resource management, geographic monitoring and so on. Due to sensor design constraints, the coverage of a single image is limited. In order to obtain wide range of remote sensing images, mosaicking technology is often used on images from different sensors and at different times. Due to the impact of sunlight incidence angle, atmospheric environment, and illumination conditions, there are radiation differences in images obtained on different platforms. The visual continuity of the mosaiced satellite images is poor, and accompanied by obvious stitching seams, which seriously affects the use of fusion images. The purpose of the color consistency processing of satellite images is to make the tone transition of the surveyed area images smooth and natural without damaging the texture information of the images and affecting the interpretation of the features.

Absolute radiometric calibration enables accurate conversion of digital number value from satellite imagery to surface reflectance [1]. However, they require parameter values of atmospheric properties at the time of image acquisition for most absolute radiometric calibration methods, which are often difficult to obtain. The method of relative radiation normalization is to make the radiation information of the target image consistent with the reference image [2], such as selecting reference images, based on overlapping regions of images, and establishing a color reference database. Absolute and relative radiometric correction methods are frequently used schemes, and the relative radiometric correction method is more widely used because the absolute radiometric correction method requires strict input conditions. However, the relative radiation normalization method also has disadvantages. Firstly, the determination standard of reference image is not uniform. Secondly, the overlap area is usually too small or most objects in the overlap area change greatly, so the regression calibration parameters are difficult to obtain. Because the transient radiation conditions of imaging are very complex, these methods can only approximate the radiation process, so mosaic images generated using these methods always have stitching seams.

The recently proposed new rendering paradigm Neural Radiance Field (NeRF) [3] is an implicit neural representation of 3D scenes and capable of self-supervised training from calibrated images. NeRF employs global MLP to regress view-dependent radiance and volume densities at any location, and can render synthesize images by applying volume from new viewpoints. Therefore, NeRF can generate realistic views of novel scenes. Currently the technique has been extended to different tasks such as relighting, scene editing and dynamic scene modeling. We address the shortcomings of traditional methods and propose a NeRF-based color consistency (NeRF-CC) method, which can change the characteristics of scenes captured in different environments in a high-quality and semantically meaningful manner.

## 2. THE PROPOSED APPROACH

Our method NeRF-CC decomposes a scene into a neural scene representation of shadows, lighting, spatial occupancy, and diffuse albedo. NeRF-CC takes as input a set of images and generates as output a 3D representation that can render images from new viewpoints under arbitrary lighting conditions. NeRF-CC can direct access to scene lighting and explicitly estimates scene intrinsic. which includes a dedicated component for fusing volumes, an essential feature for scene fusion. The pipeline of NeRF-CC is shown in Figure 1.

### 2.1 Neural Radiance Fields

NeRF defines its color $c(p)$ and density $\sigma(p)$ for any point p in 3D space. Rays are cast from the camera origin $o$ along the direction $d$ corresponding to each output pixel to render an image. The final color in image space $C(o,d)$ is obtained by integrating the color and density along the ray $(o,d)$, which can be defined as follows:

$$C(o,d) = C\left(\{p_i\}_{i=1}^{N_{depth}}\right) = \sum_{i=1}^{N_{depth}} T(t_i)\alpha(\sigma(p_i)\delta_i)c(p_i) \quad (1)$$

where $T(t_i) = \exp\left(-\sum_{j=1}^{N_{depth}-1} \sigma(p_j)\delta_j\right)$ is corresponding ray depth, $\delta_i = t_{i+1} - t_i$, $\alpha(y) = 1 - \exp(-y)$. Depth $\{t_i\}_{i=1}^{N_{depth}}$ is chosen from a uniform distribution using stratified sampling, spanning depths along $(o,d)$ which starts from the near and ends at the far camera plane. Both the color $c(p)$ and the density $\sigma(p)$ are modeled using MLP, and the final rendering is trained in a self-supervised fashion by the per-pixel color of ground truth.

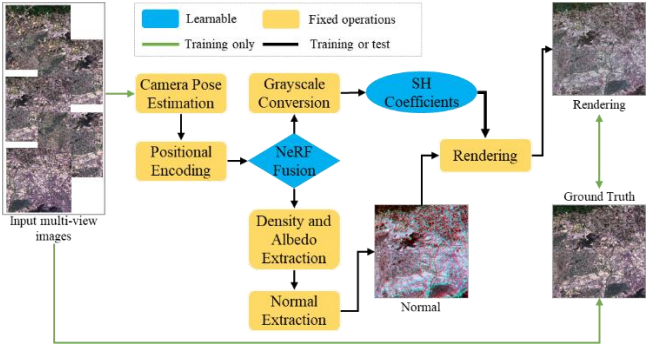

**Figure 1. The pipeline of NeRF-CC.**

While Eq. (1) allows view synthesis of high-quality free, $c(p)$ is only defined by MLP which cannot encode illumination. In other words, Eq. (1) only acquires a Lambertian model of the scene under fixed illumination. A more generalized model with view direction dependence acquires a slice of the apparent bidirectional reflectance distribution function under fixed illumination. Nevertheless, this learned representation does not contain the underlying scene semantic meaning essence, nor direct control over the illumination.

To control relighting, we introduce an explicit second-order spherical harmonic ($SH$) lighting model [4] and redefine the rendering Eq. (1) as follows:

$$C\left(\{p_i\}_{i=1}^{N_{depth}}, L\right) = A(\{p_i\}_{i=1}^{N_{depth}}) \odot Lb(N(\{p_i\}_{i=1}^{N_{depth}})) \quad (2)$$

where $\odot$ represents element-wise multiplication. $A(x)$ is the cumulative albedo color, which is produced in a similar manner to Eq. (1) by combining the output of the diffuse albedo extraction. $L$ stands for the deducible $SH$ coefficient for each image, and $b(n)$ is the $SH$ basis. $N(x)$ is the surface normal calculated from congregated ray density. To extract $N$, we distinguish the point density on the ray with respect to the raw $x, y\ and\ z$ components of the ray samples, then congregate them over all $N$ depth samples on the ray with weights $T(t_i)\alpha(\sigma(p_i)\delta_i)$, and finally normalize the output vector to unit sphere. All terms in Eq. (2) are deducible, except the regular extraction operator $N(x)$ and the $SH$ basis $b(n)$, which are based on fixed explicit models. The proposed ensemble of illumination models allows explicit re-illumination by varying $L$.

## 2.2 Volumes Fusion

To address the fusion of implicit representations of multiple images, we propose NeRFusion module (see Figure 2), which combines the advantages of TSDF-based and NeRF fusing techniques for photorealistic rendering and large-scale reconstruction. We predict the local radiance field of image sequences through direct network inference. These images are then fused together using a recurrent neural network capable of reconstructing a global sparse scene representation.

Then we aggregate features to regress a local volume $v_n$ from multiple adjacent viewpoints, representing the local radiation field. We take advantage of computing the variance and mean of per-voxel features in $\mu_i$ across adjacent viewpoints, where the variance provides abundant correspondence clues for geometric inference and the mean can contain appearance information of per-view. These two operations are also invariant to the order and number for inputs, which can handle voxels with varying numbers of visible viewpoints. A deep neural network is employed to handle mean and variance features for per-voxel to regress each viewpoint reconstruction.

$$v_t = NN([Mean_{i \in \kappa_t}\mu_i, Var_{i \in \kappa_t}\mu_i]) \quad (3)$$

where $\kappa_t$ denotes all $\kappa$ neighboring viewpoints used in each image; *Mean* and *Var* denote element-wise mean and variance operations respectively. To create scalable scene, efficient, and consistent reconstructions, we progressively fuse each frame's local feature volume $v_t$ into a global volume $v^g$ using a global volume fusion network.

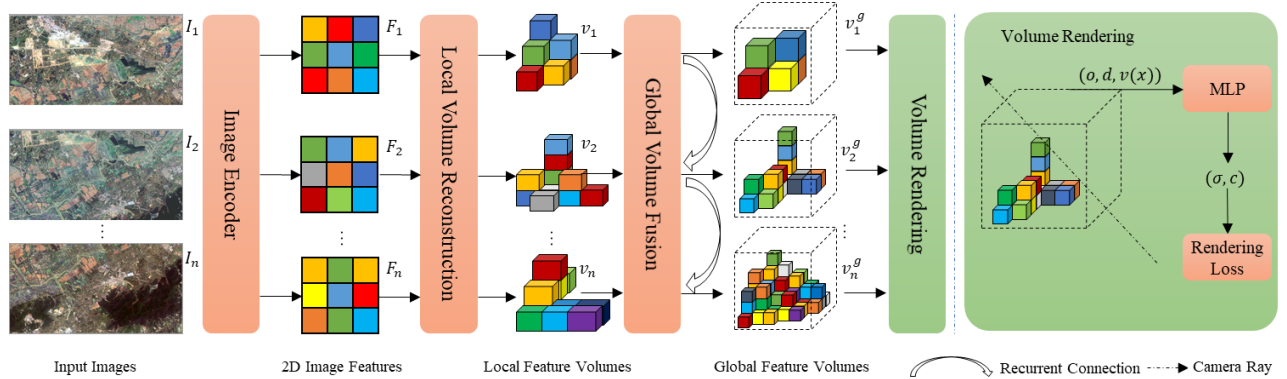

**Figure 2. The architecture of NeRFusion module.**

### 2.3 Loss Function

Our entire pipeline is trained entirely on rendering supervision from ground truth images, which does not require any additional geometric supervision. We then train the entire pipeline by a rendering loss:

$$L_{NeRF-CC} = \sum_t ||C_t - \hat{C}||_2^2 + ||C_t^g - \hat{C}||_2^2 \quad (4)$$

where $C_t$ stands for the pixel color rendered by the local reconstruction $v_t$, $\hat{C}$ stands for the pixel color of ground truth, and $C_t^g$ stands for the color rendered by the global volume $v_t^g$ after fusing images. Essentially, we use each intermediate local and global volume on each image to render new images of viewpoint and supervise them using the ground truth. Therefore, the fusion module can reasonably fuse local volumes from several input images.

## 3. EXPERIMENTS

### 3.1 Evaluation Metrics

The ground truth of remote sensing images is vital for quantitative evaluation experiments. Since it is hard to delimit what the ground truth is for a set of images with inconsistent colors, it is not feasible for this problem. Color consistency between images is a relative measure. Therefore, two metrics proposed by Xia et al. 2019 [5] will be used to evaluate the color calibration effects produced by different methods. The metrics include the color distance (CD) and the gradient loss (GL), which evaluate the color differences between overlap-corrected images and the gradient variations between the input image and the rectified image respectively. They can be defined as follows:

$$CD = \sum_{I_i \cap I_j \neq \emptyset} w_{ij} \frac{\Delta H(\hat{I}_{ij}, \hat{I}_{ji})}{N_b} \quad (5)$$

$$GL = \frac{1}{N} \sum_{i=1}^{N} \frac{\Delta G(I_i, \hat{I}_i)}{N_p} \quad (6)$$

where $I_i$ and $I_j$ are two overlapped images. $\hat{I}_i$ and $\hat{I}_j$ represent the corresponding corrected images. $\hat{I}_{ij}$ stands for the region of $\hat{I}_i$ overlapped with $\hat{I}_j$. $w_{ij}$ is the normalized weight which is proportional to the area of $\hat{I}_{ij}$. $\Delta H$ stands for the difference between the color histograms extracted from the bin, and $N_b$ is the bin number of the color histogram. $\Delta G$ stands for the difference between the gradient direction map extracted from $I_i$ and $\hat{I}_i$ by pixel, $N_p$ and is the number of valid pixels in $\hat{I}_i$. The smaller value of *CD* and *GL*, the higher the quality of the color calibration outcome.

### 3.2 Experimental Results of Different Methods

The first dataset used for testing consists of nine images acquired from the SuperView-1 satellites at urban area. The color differences between input images are relatively small in this dataset, as shown in Figure 3(a). Our method performs well on this dataset, where color differences are effectively removed between input images. We find that the left image is slightly darker than the right image for the first enlarged region by Yu et al. [6] method. For the second enlarged region by [6], we also observe that the color discrepancies between upper image and bottom image is larger than that generated by our method. There are still some tonal differences between the results of Brown et al. [7] method and Xiong et al. [8] method, especially in the lake and mountainous regions, as shown in Figure 3(b) and (c). Figure 3(d) shows the color correction results of Shen et al. [9] method, which produces the highest contrast. Especially in the right-most image area, the colors of the mountains are overly bright.

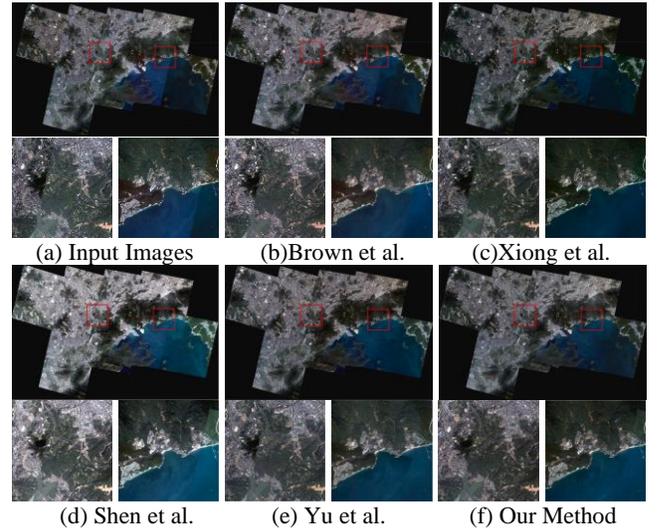

(a) Input Images     (b) Brown et al.     (c) Xiong et al.

(d) Shen et al.     (e) Yu et al.     (f) Our Method

**Figure 3. The color correction results of different methods on SuperView-1 dataset.**

The second dataset captured by UAV at village area, which consists of more than one hundred images as shown in Figure 4(a). For this dataset, the results of traditional methods have some severe problems. The results of the Brown et al. method have a consistent global tone, however, there are still some tone discrepancies between adjacent images, as shown in the enlarged regions of Figure 4(b). The Xiong et al. method produces the worst color correction results, there are still serious color differences between the corrected images. Shen et al.'s method performs well in terms of color consistency and shows only small color differences in the corrected outputs, as shown in Figure 4(d). However, the results of Shen et al.'s method suffer from serious lighting inconsistencies, and some regions are too dark (the first enlarged region). Compared to the above methods, the method of Yu et al. provides better results, but it still has some color differences. Our method provides the most visually pleasing output which is notably outperforming those produced by other methods.

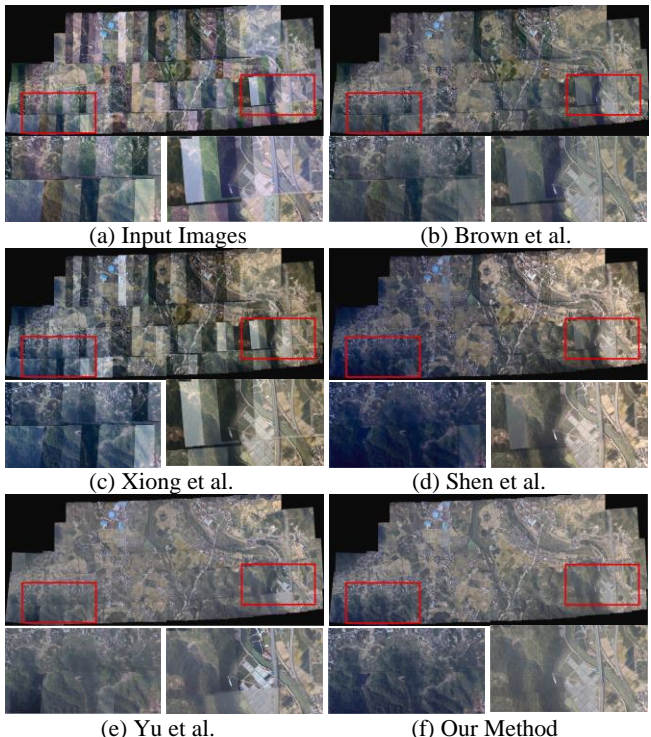

**Figure 4.** The color correction results of different methods on UAV dataset.

**Table 1** The quantitative evaluation of the color correction results produced by different methods.

|  |  | SuperView-1 | UAV |
|---|---|---|---|
| Brown et al. [7] | CD | 7.21 | 10.06 |
|  | GL | 1.79 | 1.55 |
|  | T(s) | 3.59 | 35.01 |
| Xiong et al. [8] | CD | 6.67 | 17.05 |
|  | GL | 2.15 | 1.37 |
|  | T(s) | 3.78 | 34.90 |
| Shen et al. [9] | CD | 5.82 | 9.56 |
|  | GL | 1.84 | 1.51 |
|  | T(s) | 3.64 | 43.27 |
| Yu et al. [6] | CD | 2.76 | 6.54 |
|  | GL | 1.82 | 1.59 |
|  | T(s) | 5.02 | 46.89 |
| Our Method | CD | **1.35** | **3.67** |
|  | GL | **1.62** | **1.36** |
|  | T(s) | **2.23** | **19.28** |

Besides the visual comparison, in order to convincingly illustrate the superiority of our method. The results of quality evaluation are shown in Table 1. We find that our method provides the best scores across these two datasets from Table 1, which are notably better than the rest four methods. Furthermore, we report the computation time of all methods in Table 1. We find that the computation times of Brown et al. method, Xiong et al. method, and Shen et al. method are at the same level. We also find that our method has shorter computation time than the rest four methods.

## 4. CONCLUSION

We demonstrate a method to recover replaceable neural volume representations from ambient and indirectly illuminated scene images by using visibility MLPS to approximate parts of the volume drawn integral. This paper proposes a novel neural rendering method that enables fast, high-quality, and large-scale multi-view scene reconstruction for photorealistic rendering. Our method uses a new recursive neural network to process the input multi-view images, and gradually reconstructs the global large-scale radiation field through the reconstruction and fusion of the local radiation field of each image. Compared with the traditional color consistency methods, we reconstruct the scene as a volumetric radiance field, and then obtain realistic view synthesis results.